\documentclass{article}
\usepackage{spconf,amsmath,graphicx}
\usepackage{changes}
\usepackage{mathtools}
\usepackage{float}

\usepackage[pagebackref=true,breaklinks=true,letterpaper=true,colorlinks,bookmarks=false]{hyperref}

\title{Saliency-driven Class Impressions for \\Feature Visualization of Deep Neural Networks}
\address{Author Affiliation(s)}
\name{Sravanti Addepalli \sthanks{Equal contribution author}$^{\ddagger}$ \qquad Dipesh Tamboli \footnotemark[1]$^{\dagger}$ \qquad R. Venkatesh Babu $^{\ddagger}$ \qquad Biplab Banerjee $^{\dagger}$}

\address{$^{\ddagger}$ Department of Computational and Data Sciences, Indian Institute of Science, Bengaluru, India \\
      $^{\dagger}$Indian Institute of Technology Bombay, Mumbai, India}

\begin{document}
%
\maketitle
\begin{abstract}
In this paper, we propose a data-free method of extracting \textit{Impressions} of each class from the classifier's memory. The Deep Learning regime empowers classifiers to extract distinct patterns (or features) of a given class from training data, which is the basis on which they generalize to unseen data. Before deploying these models on critical applications, it is very useful to visualize the features considered to be important for classification. Existing visualization methods develop high confidence images consisting of both background and foreground features. This makes it hard to judge what the important features of a given class are. In this work, we propose a saliency-driven approach to visualize discriminative features that are considered most important for a given task. Another drawback of existing methods is that, confidence of the generated visualizations is increased by creating multiple instances of the given class. We restrict the algorithm to develop a single object per image, which helps further in extracting features of high confidence, and also results in better visualizations. We further demonstrate the generation of \textit{negative images} as naturally fused images of two or more classes. 
\end{abstract}
\vspace{-0.2cm}
\section{Introduction}
\label{sec:intro}
Deep Learning has resulted in unprecedented progress in many of the computer vision applications such as classification \cite{krizhevsky2012imagenet}, segmentation \cite{long2015fully} and object recognition \cite{redmon2016you}. In terms of performance metrics such as classification accuracy, deep learning has outperformed the best of classical methods by a large margin \cite{krizhevsky2012imagenet}. However, one of the key issues with Deep Neural Networks is the explicability of the model. In traditional image processing algorithms, features are usually handcrafted using methods such as SIFT \cite{lowe2004distinctive} and HoG \cite{dalal2005histograms}, which are very intuitive to understand, visualize and explain. However, in a deep learning framework, features are learned by the model, and these are generated using complex nonlinear mappings from pixel space \cite{lecun1998gradient}. This makes it hard to understand features that are important for a given task. Explainable models are very important in applications such as autonomous navigation, medical diagnosis and surveillance systems. Explainability is necessary for legal compliance, identifying biases in the developed model and to improve accountability of failure cases.  In order to address these issues, there have been several works \cite{erhan2009visualizing,simonyan2013deep,zeiler2014visualizing} on the visualization of various aspects of the Deep Convolutional Networks. This includes visualization of filters, activation maps, image-specific saliency maps and visualization of the important features of a trained model \cite{simonyan2013deep,yosinski2015understanding}. Visualizing the important features of a trained model is useful to understand the inherent patterns or features that the model uses to make an inference. Such visualizations can help in validating the model and ensuring that the model does not overfit to some features that may be very specific to the domain in hand. This method can be used to test the generalizability of the model to unseen data. 

In this work, we propose a novel method of generating useful visualizations from a network, which we term as Saliency Driven Class Impressions (SCI). The proposed method uses saliency maps \cite{simonyan2013deep} to generate highly discriminative features of a given class, while suppressing weak features. Prior data-free visualization works \cite{simonyan2013deep,mopuri2018ask} focus on generating features that maximize activations of a given class. This results in the generation of weak features also, such as background, which may belong to multiple classes. Another issue with the visualizations developed by prior art is that they maximize activations by generating multiple instances of the same object. This prevents the network from developing more robust features. We use the combination of a novel saliency-driven update rule, region-growing approach and Total Variation loss \cite{rudin1992nonlinear} to develop single instances of objects. This results in the generation of more confident features, and also  more aesthetically pleasing and natural-looking visualizations of objects in a given class.

Our contribution in this work has been summarized here:
\begin{itemize}
    \item We propose a data-free method of generating visualizations containing highly discriminative features learned by the network.
    \item We demonstrate the generation of single-object images from the classifier's memory
    \item We generate natural looking \textit{negative images}, which could be used to train more robust classifiers
\end{itemize}
The organization of this paper is briefly described here. In the following section, we present related literature in the field of Visualization of Deep Networks. Section-\ref{sec:proposed method} explains our proposed method in detail. We conclude the paper with our analysis in Section-\ref{sec:results}.

\section{Related Literature}
\label{sec:format}
\subsection{Review of Visualization methods} 
The first layer filters of Convolutional Neural Networks can be directly visualized in pixel level. However, we need more sophisticated methods to understand the dynamics of the remaining layers. One popular line of work in this area involves generation of images by maximizing activations at any given intermediate layer, or at the final output of the network \cite{erhan2009visualizing,Le:2012:BHF:3042573.3042641,simonyan2013deep,mopuri2018ask}.
\subsection{Image-Specific Class Saliency Maps}
A method for class saliency visualization was proposed by Simonyan et al. \cite{simonyan2013deep}. The authors define Saliency maps as the \textit{spatial support} of a class in a given image. Saliency maps highlight regions that contribute to the classification of a given image to its respective class. This is very useful to visually understand the features that are important for classifying the object. Saliency maps are computed by taking derivative of the class scores with respect to the input image. The magnitude of the derivative indicates the pixels that need to be modified the least to affect the class score the most. Hence, saliency maps can be used to judge the importance of each pixel. Saliency based methods such as CAM \cite{zhou2016learning}, Grad-CAM \cite{selvaraju2017grad} and Grad-CAM++ \cite{chattopadhay2018grad} have emerged as popular tools to highlight features in a given input image. In this work, we use saliency maps to define an adaptive learning rate for updating each pixel in the generated visualizations.
\subsection{Class Impressions}
\label{CI}
The method of generating visualizations from a trained classifier was first proposed by Erhan et al. \cite{erhan2009visualizing}, which was applied on Convolutional Networks later by Simonyan et al. \cite{simonyan2013deep}. There are several works on the addition of regularizers to improve the quality of generated images \cite{olah2017feature}. We use the recent methodology proposed by Mopuri et al. \cite{mopuri2018ask} as a starting point for this work. They initially start with a noise image and iteratively update this using gradient ascent to maximize the logits of a given class. A set of transformations such as random rotation, scaling, RGB jittering and random cropping between iterations ensures that the generated images are robust to these transformations; a feature that natural images typically possess. The generated images are termed as \textit{Class Impressions} (Fig.\ref{fig:TV}(a)). Some of the drawbacks of this approach are as follows: Class Impressions lack the texture and smoothness of natural images, logits are maximized by creating multiple instances of the same object, and this approach does not necessarily generate only the discriminative features of a given class. As seen in Fig.\ref{fig:TV}(a), generated images contain background and foreground features. Background features are clearly not discriminative features, as they may be common to many other classes as well. 

We address these issues in the proposed approach, which is described in detail in the following section.

\section{Proposed Method}
\label{sec:proposed method}
In this section, we describe each of the improvements proposed in this paper, followed by the combined impact of all. We further discuss the methodology developed for generation of single-object images, followed by a brief note on the hyper-parameters used in the loss function. Finally, we demonstrate generation of natural-looking \textit{negative images}.

The notation used in the remainder of this paper is as follows: $SCI _{c}$ denotes the Saliency-driven Class Impressions generated for class $c$. $L _{c}(I)$ denotes the pre-softmax output corresponding to class $c$ for an image $I$. $I _{i, j}$ denotes the intensity of the pixel at the location $ (i, j)$ in the image $I$.

A VGG-F network \cite{chatfield2014return} pre-trained on ILSVRC \cite{russakovsky2015imagenet} dataset is used for all the experiments described in this paper. The generated Class Impressions are of dimension $224 \times 224 \times 3$, which is the same as the dimension of images in the ILSVRC dataset. We start with an initial mean image, and add a noise image with each pixel sampled independently from $U[0 ,255]$. The image is updated iteratively for a fixed number of iterations using gradient ascent to maximize a pre-defined loss, described later in this section. A set of random transformations \cite{mopuri2018ask} (Section-\ref{CI}) are applied after every iteration. We also control the gradient flow to constraint the Class Impressions (CIs) to develop a single object only. 
\subsection{Total Variation Loss as a Natural Image prior}
One of the key statistical properties that characterize images, specifically natural images, is their spatial smoothness. In order to enforce this prior, the objective function of generating Class Impressions \cite{mopuri2018ask} is modified to include the Total Variation (TV) loss \cite{rudin1992nonlinear}, as shown below:
\begin{equation}
\label{eq1}
SCI _{c} = \underset{I}{\mathrm{argmax}}\;L_{c}(I) - \lambda _{1} \sum_{i, j}|I _{i+1, j} - I _{i, j}| + |I _{i, j+1} - I _{i, j}|
\end{equation}

\begin{figure}
  \centering
  \fbox{\includegraphics[width=0.95\linewidth]{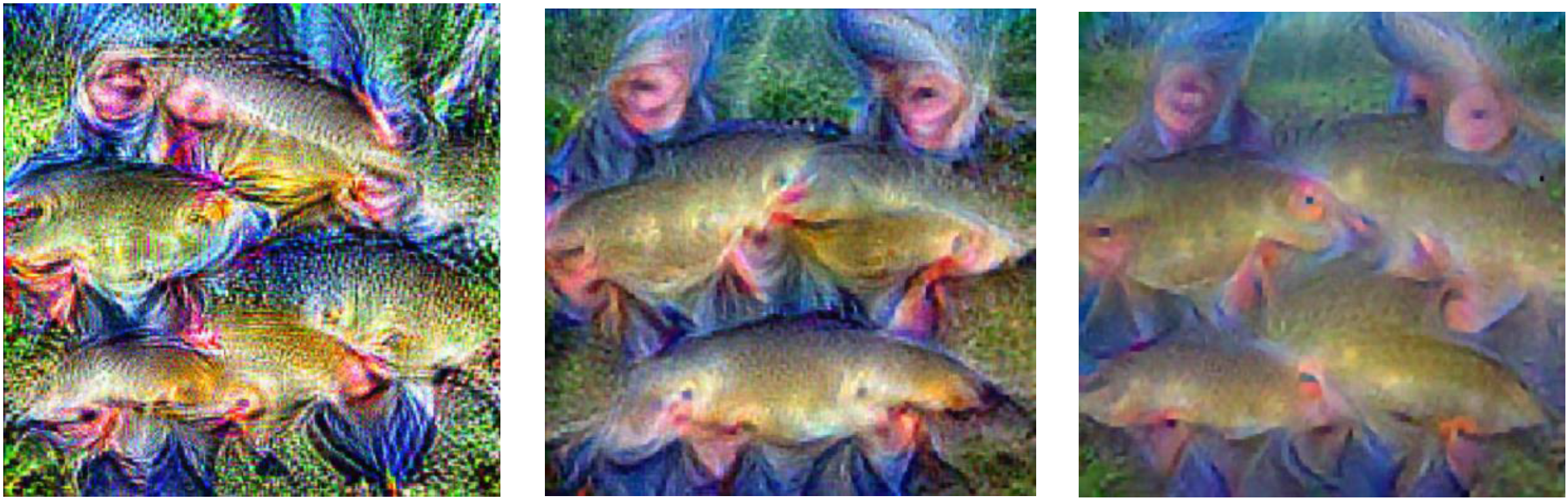}}
  \caption{Class \textit{Tench}: (a) CI (b, c) CI with TV loss} 
  \label{fig:TV}
\end{figure}

Here, $\lambda _{1}$ is the weightage given to the TV loss component. The images generated using this loss are shown in Fig.\ref{fig:TV}(b) and (c). This loss penalizes the high frequency components, thereby producing locally smooth images. After using TV loss, texture of the fish resembles that of the original image better. It must be noted that this process is not equivalent to applying a low pass filter (LPF) on the output image. In the case of applying LPF to the output image, important edge information is also lost. However, when TV loss is used during training, the edge information that is crucial for classification is preserved, as it helps maximize activations of that class. For example, eye of the fish is more predominant in Fig.\ref{fig:TV}(b) and (c) when compared to Fig.\ref{fig:TV}(a). The optimization of each component of loss function in Eq.\ref{eq1} is done individually. The TV loss is applied only once in every $k$ iterations. For Fig.\ref{fig:TV}(b), $k$ $=$ $2$, and for Fig.\ref{fig:TV}(c), $k$ $=$ $1$.
\begin{figure}
  \centering
  \fbox{\includegraphics[width=0.95\linewidth]{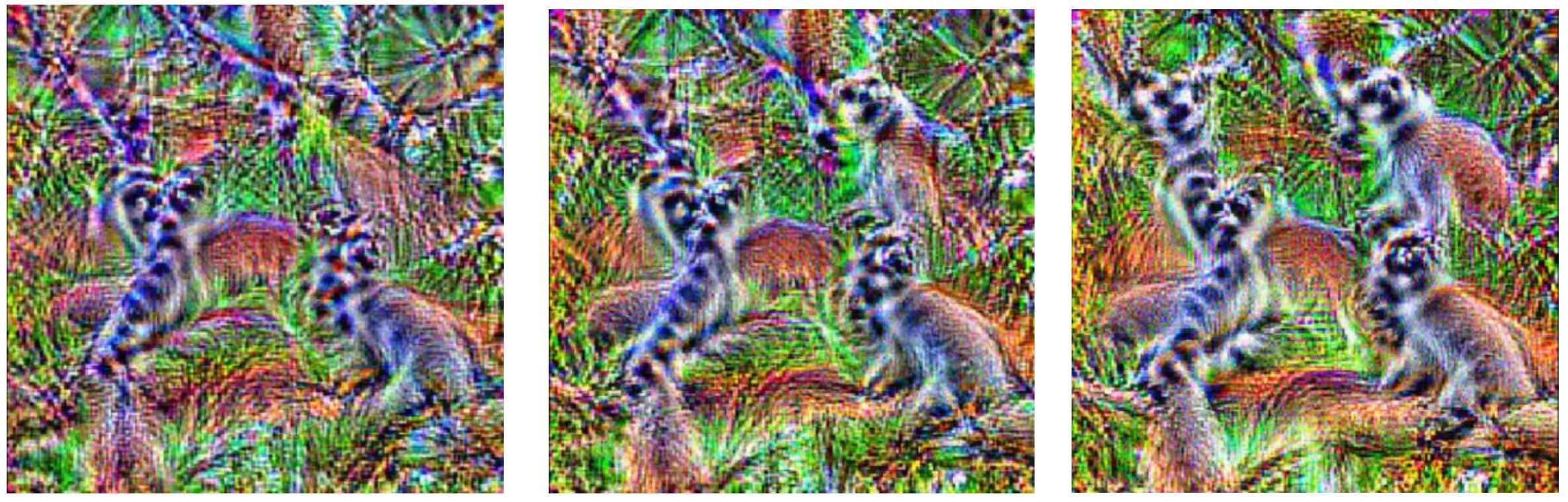}}
  \caption{\textit{Madagascar Cat}: Generated CIs after (a) 100, (b) 200, and (c) 500 iterations} 
  \label{fig:CI}
\end{figure}

\subsection{Saliency-driven Class Impressions}
\label{subsec:sdci}
Traditionally, Class Impressions (CIs) \cite{mopuri2018ask} are generated by maximization of the pre-softmax value of a given class. This can be achieved in two ways:
\begin{itemize}
    \item By improving the confidence of generated features in a small subset of pixels
    \item By maximizing the number of pixels that have the required features
\end{itemize}
A study of the CIs generated over iterations shows that the optimization follows a mix of both approaches. This results in images that have repeated structures. Fig.\ref{fig:CI} shows the generated class impressions for the class, Madagascar Cat. After 100 iterations there are 2 cats clearly visible in the image. This number increases to 3 after 200 iterations and 4 after 500 iterations. Contrary to this, we expect the initial features to become more discriminative with progress in the number of iterations. This could potentially be achieved by adding a loss that penalizes the number of pixels that contribute to the correct classification of an image. However, the drawback of this approach is that a threshold needs to be selected to count only some of the pixels as \textit{strong contributors}. This threshold needs to be image-specific, as the magnitude of gradient and number of contributing pixels can be different for different images, based on their scale and texture. 

To address this issue, we propose to control the learning rate of each pixel adaptively, based on the Saliency maps computed at each step. This ensures that the pixels that maximally contribute to the objective function develop at a faster rate when compared to the non-contributing pixels. Generation of Saliency maps does not add to the computational complexity, as the derivative of the loss function with respect to the input is also required to be computed for the generation of CIs. We start with a uniform learning rate for all pixels. The learning rate update rule is illustrated here. We first calculate $grad_{cum,i}$, which is the cumulative weighted gradient upto the $i^{th}$ iteration as shown below: \\
\begin{equation}{\small
  grad_{cum, i} = \left \{
  \label{eq4}
  \begin{aligned}
    grad_{cum, i-1} \cdot i + C_{1,i} \cdot \frac{grad_{i}}{\|grad_{i}\|}, ~\text{if}\ i< t\:\:\:\:\:\:\:\: \:\:\:\:\\
    grad_{cum, i-1} \cdot  i+ C_{2} \cdot  \frac{grad_{i}}{\|grad_{i}\|}, \:\: \text{otherwise}\:\:\:\:\:\:\:\:\:
  \end{aligned} \right.}
\end{equation}
The weightage given to the direction of current gradient $grad_{i}$ is $C_{1,i}$ upto the $t^{th}$ iteration and $C_2$ beyond this. The value of $C_{1,i}$ is ramped up from $0$ to $C_2$ over $t$ iterations as shown below:
\begin{equation}
    C_{1,i} = C_2\cdot \frac{i}{t}
\end{equation}
The value of $C_2$ is chosen as $4$ and $t$ is set to 150 in all our experiments. The cumulative weighted gradient, $grad_{cum,i}$ is normalized to obtain the final learning rate map $lr_{map,i}$ at every iteration.
\begin{equation}
\label{eq5}
 lr_{map,i} = \frac{grad_{cum,i}}{\|grad_{cum,i}\|} 
\end{equation}
$lr_{map,i-1}$ is used as the final pixel-wise learning rate for the $i^{th}$ iteration. 
\subsection{Methodology for Single object generation}
\begin{figure}[htb]
\begin{minipage}[b]{.48\linewidth}
  \centering
  \centerline{\includegraphics[width=4.0cm]{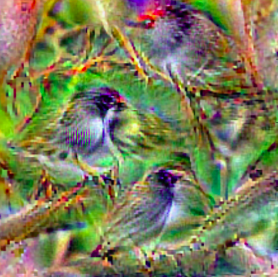}}
  \centerline{pre-Class impressions}\medskip
\end{minipage}
\hfill
\begin{minipage}[b]{0.48\linewidth}
  \centering
  \centerline{\includegraphics[width=4.0cm]{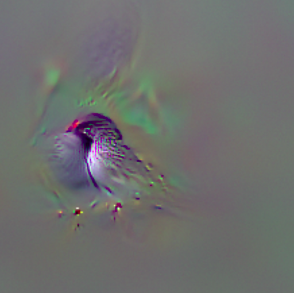}}
  \centerline{post-Class Impressions}\medskip
\end{minipage}
\vspace{-0.6cm}
\caption{Pre-CI and the corresponding post-CI}
\label{fig:pre-CI}
\end{figure}

The Saliency driven approach described in the above section helps in developing the important regions of an image faster than other regions. In order to develop a single object for a given image, it is also important to ensure that the salient regions are connected. In this section, we describe the methodology used to achieve this. We first develop initial \textit{Pre-Class Impressions (pre-CI)} using the process described in Section-\ref{subsec:sdci}. With this setting, Class Impressions are developed for $500$ iterations. Next, we select the most activated region of the CIs based on their $lr_{map}$ values and develop this region again, as shown in Fig. \ref{fig:pre-CI}.

The methodology used for selecting the most activated region is described here. Let $circ(x,y,r)$ denote an image of the same size as that of the pre-CI, with a value of $1$ within a circle of radius $r$ centered at $(x,y)$, and a value of $0$ otherwise. We define the region around $(x,y)$ as the most activated region, if the circle of radius $r$ around this pixel has the highest magnitude of learning rate as shown below:

\begin{equation}
\label{eq3} {
 (x,y)_{max} = \underset{(x,y)}{\mathrm{argmax}}\; \sum_{i}\sum_{j}|lr_{map} \cdot circ(x,y,r)|}
\end{equation}

Here, $lr_{map}$ corresponds to the pixel-wise learning rate computed using Eq.\ref{eq5} for the final iteration. A circular mask, $ circ((x,y)_{max},r)$ is constructed to be centered at the most activated region $(x,y)_{max}$. This is multiplied with the $lr_{map}$ in Eq.\ref{eq5} to generate the final learning rate map for each pixel. The mask helps in developing the most activated region first, and further expanding this into a full object. The radius of this circular mask starts with 1 and increases linearly over $150$ iterations to a value of $150$, beyond which it is kept constant. 

The inclusion of Total Variation loss ensures that the region around the object fades out, and only the high confidence features sustain. 

\section{Results and Conclusions}
\label{sec:results}
\begin{figure}[htb]
\begin{minipage}[b]{0.48\linewidth}
  \centering
  \centerline{\includegraphics[width=4.0cm]{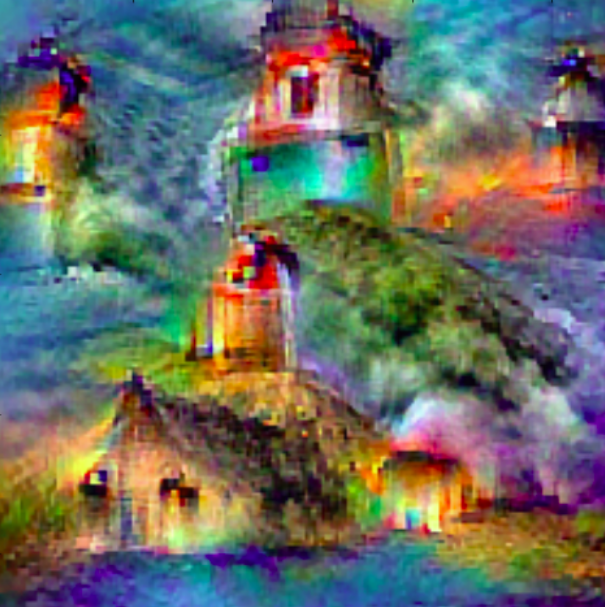}}
  \centerline{Watch-Tower, AAA \cite{mopuri2018ask}}\medskip
\end{minipage}
\hfill
\begin{minipage}[b]{0.48\linewidth}
  \centering
  \centerline{\includegraphics[width=4.0cm]{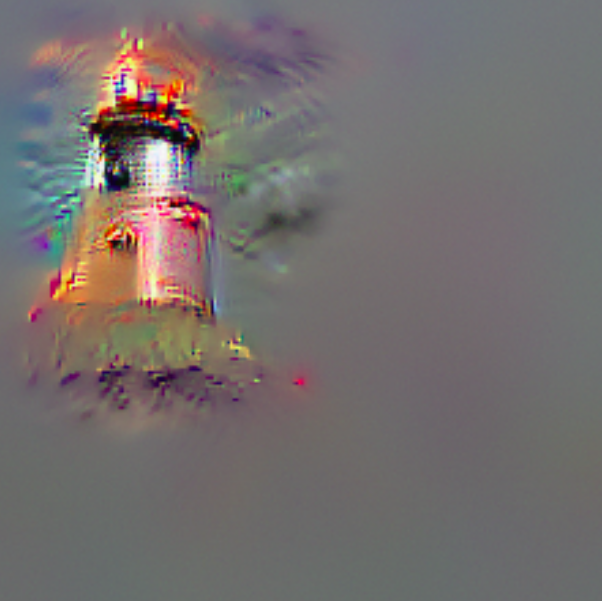}}
  \centerline{Watch-Tower, Ours}\medskip
\end{minipage}
%

\begin{minipage}[b]{0.48\linewidth}
  \centering
  \centerline{\includegraphics[width=4.0cm]{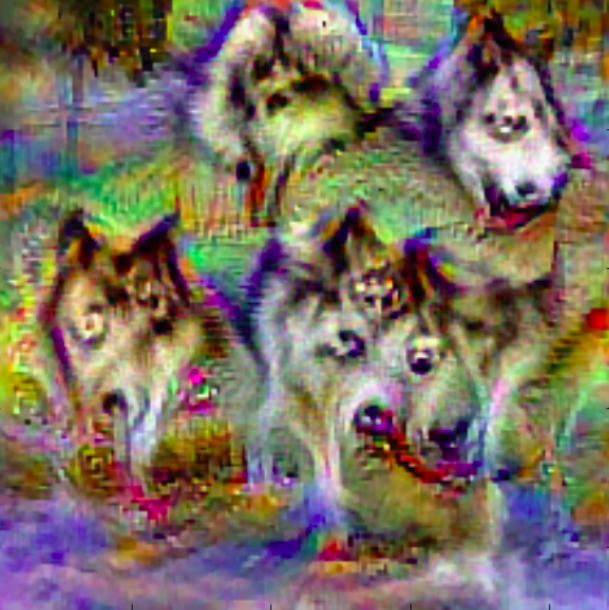}}
  \centerline{Fox, AAA \cite{mopuri2018ask}}\medskip
\end{minipage}
\hfill
\begin{minipage}[b]{0.48\linewidth}
  \centering
  \centerline{\includegraphics[width=4.0cm]{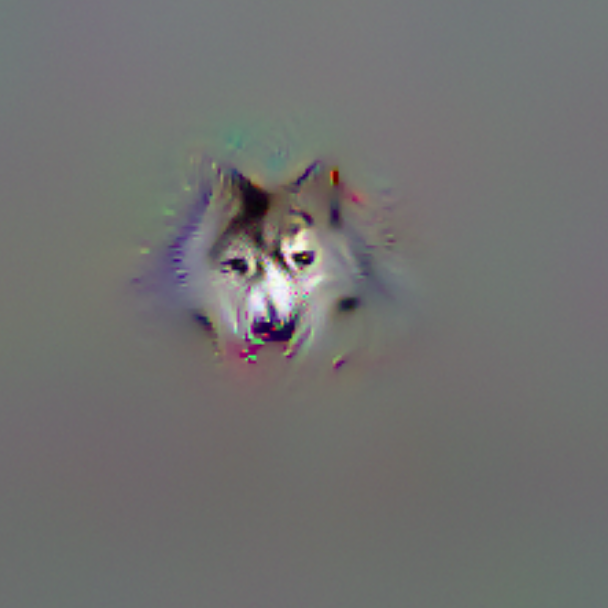}}
  \centerline{Fox, Ours}\medskip
\end{minipage}
\begin{minipage}[b]{0.48\linewidth}
  \centering
  \centerline{\includegraphics[width=4.0cm]{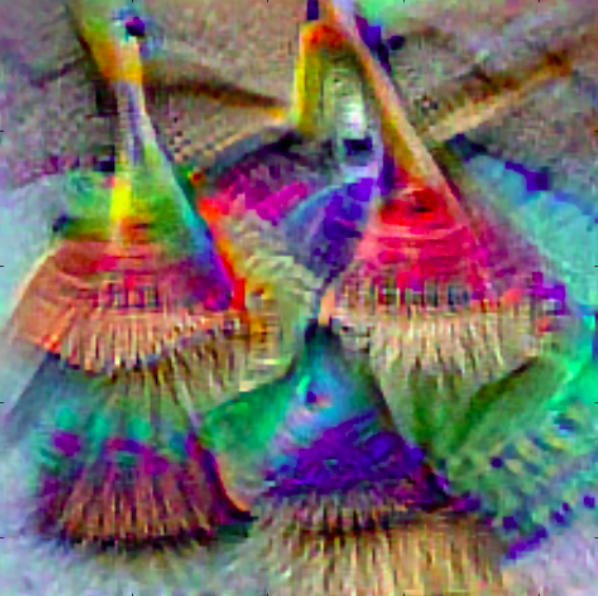}}
  \centerline{Broom, AAA \cite{mopuri2018ask}}\medskip
\end{minipage}
\hfill
\begin{minipage}[b]{0.48\linewidth}
  \centering
  \centerline{\includegraphics[width=4.0cm]{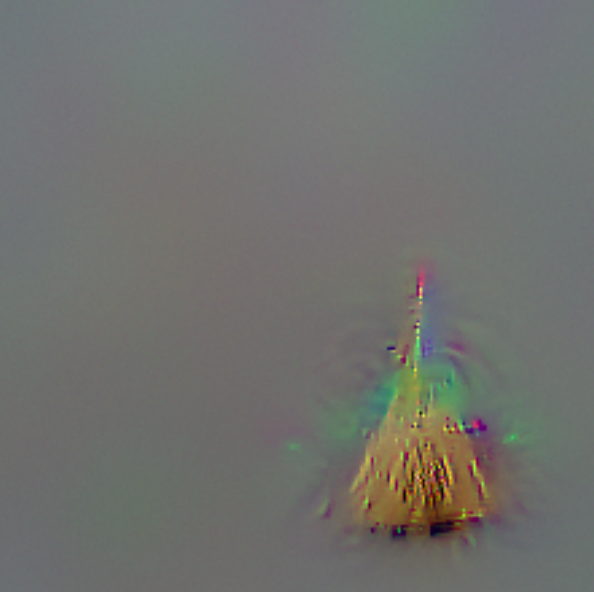}}
  \centerline{Broom, Ours}\medskip
\end{minipage}
\vspace{-0.5cm}
\caption{Comparison with existing methods}
\label{fig:compare}
\end{figure}

We show a comparison of our results with that of the baseline by Mopuri et al. \cite{mopuri2018ask} in Fig. \ref{fig:compare}. It can be seen that the proposed method generates images of significantly improved quality, containing single instances of objects, unlike the corresponding results of Ask, Aquire and Attack (AAA) \cite{mopuri2018ask}, where multiple instances are developed.

\subsection{Images developed by fusion of two classes}
\begin{figure}[H]
\begin{minipage}[b]{.48\linewidth}
  \centering
  \centerline{\includegraphics[width=4.0cm]{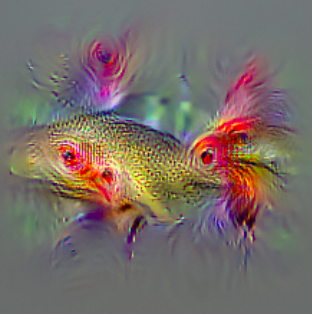}}
  \centerline{Fusion of fish and hen}\medskip
\end{minipage}
\hfill
\begin{minipage}[b]{0.48\linewidth}
  \centering
  \centerline{\includegraphics[width=4.0cm]{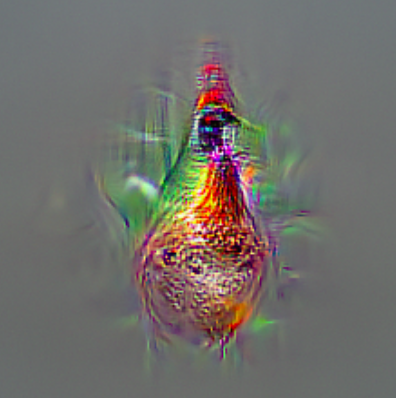}}
  \centerline{Fusion of bottle and hen}\medskip
\end{minipage}
\vspace{-0.5cm}
\caption{Negative Images}
\label{fig:neg_img}
\end{figure}

We use the proposed method to generate negative class images by fusing class impressions of 2 or more classes. Here, we start from a single pixel and develop CIs of one class around this pixel for a few iterations. Next, the second class is developed from a different region for a few iterations. This generates a natural fusion between the class impressions of both classes, as can be seen in Fig.\ref{fig:neg_img}. These negative class images do not belong to either of the two classes confidently,  

\noindent as they contain features that belong to the other class. Such images can be used to train robust classifiers, improve the confidence predictions of a classifier and also for developing a reject option for out-of-distribution samples.

\pagebreak





\bibliographystyle{IEEEbib}
\bibliography{refs}
\end{document}